\PassOptionsToPackage{unicode}{hyperref}
\PassOptionsToPackage{hyphens}{url}
\documentclass[
]{article}
\usepackage{amsmath,amssymb}
\usepackage{iftex}
\ifPDFTeX
  \usepackage[T1]{fontenc}
  \usepackage[utf8]{inputenc}
  \usepackage{textcomp} 
\else 
  \usepackage{unicode-math} 
  \defaultfontfeatures{Scale=MatchLowercase}
  \defaultfontfeatures[\rmfamily]{Ligatures=TeX,Scale=1}
\fi
\usepackage{lmodern}
\ifPDFTeX\else
\fi
\IfFileExists{upquote.sty}{\usepackage{upquote}}{}
\IfFileExists{microtype.sty}{
  \usepackage[]{microtype}
  \UseMicrotypeSet[protrusion]{basicmath} 
}{}
\makeatletter
\@ifundefined{KOMAClassName}{
  \IfFileExists{parskip.sty}{%
    \usepackage{parskip}
  }{
    \setlength{\parindent}{0pt}
    \setlength{\parskip}{6pt plus 2pt minus 1pt}}
}{
  \KOMAoptions{parskip=half}}
\makeatother
\usepackage{xcolor}
\usepackage{color}
\usepackage{fancyvrb}

\DefineVerbatimEnvironment{Highlighting}{Verbatim}{commandchars=\\\{\}}
\newenvironment{Shaded}{}{}

\newcommand{\DataTypeTok}[1]{\textcolor[rgb]{0.56,0.13,0.00}{#1}}
\newcommand{\DecValTok}[1]{\textcolor[rgb]{0.25,0.63,0.44}{#1}}

\newcommand{\FunctionTok}[1]{\textcolor[rgb]{0.02,0.16,0.49}{#1}}

\newcommand{\KeywordTok}[1]{\textcolor[rgb]{0.00,0.44,0.13}{\textbf{#1}}}

\newcommand{\OtherTok}[1]{\textcolor[rgb]{0.00,0.44,0.13}{#1}}

\newcommand{\StringTok}[1]{\textcolor[rgb]{0.25,0.44,0.63}{#1}}

\usepackage{longtable,booktabs,array}
\usepackage{calc} 
\usepackage{etoolbox}
\makeatletter
\patchcmd\longtable{\par}{\if@noskipsec\mbox{}\fi\par}{}{}
\makeatother
\IfFileExists{footnotehyper.sty}{\usepackage{footnotehyper}}{\usepackage{footnote}}
\makesavenoteenv{longtable}
\providecommand{\tightlist}{%
  \setlength{\itemsep}{0pt}\setlength{\parskip}{0pt}}
\ifLuaTeX
  \usepackage{selnolig}  
\fi
\IfFileExists{bookmark.sty}{\usepackage{bookmark}}{\usepackage{hyperref}}
\IfFileExists{xurl.sty}{\usepackage{xurl}}{} 
\hypersetup{
  hidelinks,
  pdfcreator={LaTeX via pandoc}}

\author{}
\date{}

\begin{document}

\hypertarget{model-routing-as-a-trust-problem-route-receipts-for-adaptive-ai-systems}{%
\section{Model Routing as a Trust Problem: Route Receipts for Adaptive
AI
Systems}\label{model-routing-as-a-trust-problem-route-receipts-for-adaptive-ai-systems}}

\textbf{Vincent Schmalbach}\\
Independent Researcher\\
https://www.vincentschmalbach.com/\\
vschmalbach@vschmalbach.com

\hypertarget{abstract}{%
\subsection{Abstract}\label{abstract}}

AI products often route requests through version aliases, service tiers,
tool choices, regional endpoints, fallback rules, or safety handling
before responding. These routing steps are documented product surfaces
in several widely used AI platforms and serving stacks.

Routing helps AI services stay affordable, fast, and available at scale,
and it shapes trust. Trust can break when routing changes the cost,
quality, or accountability of a response without the user being able to
tell what happened. ``Which model answered?'' is only part of the audit
question. The runtime path matters.

Adaptive AI systems should produce a runtime transparency artifact
called the \textbf{route receipt}. A route receipt is a compact record
of the route that served a request. It should capture enough material
facts for people relying on the output to reconstruct important routing
decisions without exposing proprietary internals or hidden reasoning.

Route transparency should be part of model documentation. Model cards
describe trained model artifacts, while route receipts describe the
runtime conditions under which a particular answer was produced. The
paper introduces the route-receipt concept, a minimal schema and
redaction model, and a documentation-based survey of selected platforms
showing that receipt fragments already exist without a portable
per-answer record.

\begin{center}\rule{0.5\linewidth}{0.5pt}\end{center}

\hypertarget{missing-route-records-in-ai}{%
\subsection{1. Missing route records in
AI}\label{missing-route-records-in-ai}}

A team ships an AI feature to production, then a customer reports an
incorrect answer. The error can be a contract exception omitted from the
answer, a support policy that came back wrong, a research answer that
cited stale material, or an internal agent that timed out and returned a
partial result.

An engineer reviews the audit logs.

The audit log records the prompt, response, and token counts. It may
also include a request ID, a timestamp, and a model name. In mature
systems, it may also record latency metrics, retrieved document IDs, and
billing records.

Which path returned this answer?

Was the request run on the model the team expected, or on an alias that
had changed since the test run? Did it go through priority capacity or
standard capacity, use a fallback after the first route failed, use web
search, file retrieval, cached context, or no tool at all, or get
processed inside the expected data zone or through a global endpoint
chosen for availability?

In many AI systems, those facts are split across different operational
systems or unavailable entirely, and only the provider can see them.
Users see a response, developers see a model name, administrators see
usage totals, and the user often does not receive a compact, portable
account of the service path.

Without that account, post-hoc review begins with guesses about the
route.

Public discussion of AI still often treats the model as the unit of
explanation, the thing that answered, hallucinated, refused, or was
upgraded. That language made sense when the product story was organized
around a single named model. That framing no longer matches how modern
AI services work. Modern AI services increasingly make runtime decisions
about model version, tool access, compute budget, service tier,
processing location, safety path, and fallback behavior.

Those choices can materially shape the answer even when users never see
them. A different route can change the output while the product label
stays the same.

Adaptive AI systems should produce route receipts: compact runtime
records of serving conditions. A route receipt records transaction facts
without exposing reasoning or router source code.

Model cards and route receipts answer different questions. A model card
describes a trained model or model family, including intended uses,
evaluation results, limitations, and known risks. A route receipt
describes the runtime path of one answer: which version or route served
it, what tier or effort was used, whether tools or fallbacks were
involved, and what route facts were redacted. Model cards document
design time; route receipts document runtime.

Most casual users will never inspect this metadata, so the design should
make it easy to ignore.

\begin{center}\rule{0.5\linewidth}{0.5pt}\end{center}

\hypertarget{routing-infrastructure}{%
\subsection{2. Routing infrastructure}\label{routing-infrastructure}}

This section looks at routing behavior in selected platforms and serving
stacks.

OpenAI's Priority processing documentation says response objects include
the service tier used to process a request. If traffic ramps too
quickly, some Priority requests may be downgraded to Standard, billed at
Standard rates, and shown as \texttt{service\_tier="default"} in the
response. The requested and effective routes can diverge, and the
response carries a receipt-like field.
(\href{https://developers.openai.com/api/docs/guides/priority-processing}{OpenAI
API Documentation})

Anthropic's service-tier documentation describes Priority, Standard, and
Batch tiers, says \texttt{service\_tier="auto"} can use Priority Tier
when available and fall back to other capacity, and says the response
\texttt{usage} object includes the assigned service tier. Priority
commitments use a specific model version.
(\href{https://platform.claude.com/docs/en/api/service-tiers}{Anthropic
API Documentation})

AWS Bedrock documents the same routing behavior in its service-tier
documentation. It describes Reserved, Priority, Standard, and Flex
tiers, says the service tier configuration for a served request is
visible in the API response and AWS CloudTrail events, and notes that
CloudWatch metrics include \texttt{ResolvedServiceTier}, which shows the
tier that served each request.
(\href{https://docs.aws.amazon.com/bedrock/latest/userguide/service-tiers-inference.html}{AWS
Documentation})

Versioning constrains routing. Google's Gemini documentation
distinguishes stable, preview, latest, and experimental model
identifiers. Stable models usually do not change, and Google says most
production apps should use a specific stable model. A \texttt{latest}
alias points to the latest release of a model variation and is
hot-swapped when a new release appears, with notice before the version
behind \texttt{latest} changes.
(\href{https://ai.google.dev/gemini-api/docs/models}{Google AI for
Developers}) Microsoft Foundry's lifecycle documentation treats model
versions as operationally significant. Models move through lifecycle
stages from preview to retirement, and customers are expected to
evaluate replacements and migrate workloads.
(\href{https://learn.microsoft.com/en-us/azure/foundry/openai/concepts/model-retirements}{Microsoft
Learn})

Some providers build routing into the product. AWS Bedrock's intelligent
prompt routing analyzes an incoming prompt, predicts response quality
for models in a selected family, chooses a model based on quality and
cost, and returns the model used. The same documentation notes that
routing may not always be optimal for specialized cases.
(\href{https://docs.aws.amazon.com/bedrock/latest/userguide/prompt-routing.html}{AWS
Documentation}) Microsoft Foundry's model router analyzes prompts in
real time and routes them to eligible models according to settings such
as Balanced, Cost, or Quality mode. Its documentation also warns that
the effective context window is limited by the smallest underlying
model.
(\href{https://learn.microsoft.com/en-us/azure/foundry/openai/concepts/model-router}{Microsoft
Learn})

OpenAI documents web search routing. OpenAI's web-search documentation
says models can search the web before generating a response, provide
sourced citations, and decide whether to search based on the input when
web search is enabled in the Responses API. Responses that use the
web-search tool include a \texttt{web\_search\_call} output item and
inline URL citations.
(\href{https://developers.openai.com/api/docs/guides/tools-web-search}{OpenAI
API Documentation})

Region and provider routing is core platform infrastructure. OpenAI's
data-controls documentation describes project-level data residency,
regional domain prefixes, and regional processing support for some
regions and services.
(\href{https://developers.openai.com/api/docs/guides/your-data}{OpenAI
API Documentation}) Vertex AI documentation says global endpoints can
improve availability and reduce resource-exhausted errors. It also warns
against using the global endpoint when ML processing location
requirements apply, because the user cannot control which region
processes the request.
(\href{https://docs.cloud.google.com/vertex-ai/generative-ai/docs/learn/locations}{Google
Cloud Vertex AI Documentation}) OpenRouter documents provider routing
that prioritizes price by default, takes recent outages into account,
uses remaining providers as fallbacks, and lets users sort providers by
price, throughput, or latency.
(\href{https://openrouter.ai/docs/guides/routing/provider-selection}{OpenRouter})
It also documents zero-data-retention controls that limit routing to
endpoints with a ZDR policy.
(\href{https://openrouter.ai/docs/guides/features/zdr}{OpenRouter})

These examples show routing as ordinary platform behavior and an
operational surface of AI products.

\hypertarget{route-observability-surfaces-in-vendor-docs}{%
\subsection{2.1 Route-observability surfaces in vendor
docs}\label{route-observability-surfaces-in-vendor-docs}}

The vendor documentation cited above shows which routing surfaces are
documented. It works best as a structured comparison of which parts of a
proposed route receipt are already exposed, partly exposed, or not
documented as exposed across selected inference platforms and serving
stacks.

The comparison is illustrative rather than exhaustive, and the included
platforms are those whose public documentation exposes at least one
routing-relevant surface. It uses five status labels: \textbf{exposed in
API}, \textbf{exposed in logs/dashboard}, \textbf{partially exposed},
\textbf{not documented}, and \textbf{not applicable}. All documentation
cited in the table was accessed on April 29, 2026; that date applies to
each cited cell. ``Not documented'' means not documented in the surveyed
source as a response, log, dashboard, or export field that can function
as a per-answer receipt; it does not mean the provider lacks adjacent
policy, contract, or admin controls. This is a narrower comparison than
a benchmark. Current platform behavior includes receipt fragments, but
no surveyed platform exposes a portable per-answer receipt for the full
route.

\begin{longtable}[]{@{}
  >{\raggedright\arraybackslash}p{(\columnwidth - 4\tabcolsep) * \real{0.3333}}
  >{\raggedright\arraybackslash}p{(\columnwidth - 4\tabcolsep) * \real{0.3333}}
  >{\raggedright\arraybackslash}p{(\columnwidth - 4\tabcolsep) * \real{0.3333}}@{}}
\caption{Route-observability surfaces documented for selected AI
inference platforms and serving stacks as of April 29, 2026. Compact
labels keep the table readable, and the citations identify the surveyed
documentation for each row.}\tabularnewline
\toprule\noalign{}
\begin{minipage}[b]{\linewidth}\raggedright
Platform
\end{minipage} & \begin{minipage}[b]{\linewidth}\raggedright
Documented receipt fragments
\end{minipage} & \begin{minipage}[b]{\linewidth}\raggedright
Gaps in surveyed docs
\end{minipage} \\
\midrule\noalign{}
\endfirsthead
\toprule\noalign{}
\begin{minipage}[b]{\linewidth}\raggedright
Platform
\end{minipage} & \begin{minipage}[b]{\linewidth}\raggedright
Documented receipt fragments
\end{minipage} & \begin{minipage}[b]{\linewidth}\raggedright
Gaps in surveyed docs
\end{minipage} \\
\midrule\noalign{}
\endhead
\bottomrule\noalign{}
\endlastfoot
OpenAI API & Service tier, web-search traces, citations, data-residency
configuration.
\href{https://developers.openai.com/api/docs/guides/priority-processing}{1},
\href{https://developers.openai.com/api/docs/guides/tools-web-search}{8},
\href{https://developers.openai.com/api/docs/guides/your-data}{43} & No
separate alias-resolution receipt or broad fallback record. \\
Anthropic API & Assigned service tier and web-search tool blocks, result
blocks, citations, and usage counts.
\href{https://platform.claude.com/docs/en/api/service-tiers}{2},
\href{https://platform.claude.com/docs/en/agents-and-tools/tool-use/web-search-tool}{44}
& No general resolved-version or full fallback-path receipt. \\
Amazon Bedrock & Service tier in API, CloudTrail, and metrics; selected
model for prompt routing.
\href{https://docs.aws.amazon.com/bedrock/latest/userguide/service-tiers-inference.html}{3},
\href{https://docs.aws.amazon.com/bedrock/latest/userguide/prompt-routing.html}{6}
& No surveyed tool-search trace or per-request region receipt. \\
Google Vertex AI / Gemini & Stable/latest model identifiers, thinking
budget, routing configuration, and global endpoint behavior.
\href{https://ai.google.dev/gemini-api/docs/models}{4},
\href{https://docs.cloud.google.com/vertex-ai/generative-ai/docs/learn/locations}{9},
\href{https://docs.cloud.google.com/vertex-ai/generative-ai/docs/reference/rest/v1/GenerationConfig}{18}
& No served-route receipt in the surveyed docs. \\
Microsoft Foundry & Lifecycle notices and router modes such as Balanced,
Cost, and Quality.
\href{https://learn.microsoft.com/en-us/azure/foundry/openai/concepts/model-retirements}{5},
\href{https://learn.microsoft.com/en-us/azure/foundry/openai/concepts/model-router}{7}
& No documented per-answer route record. \\
OpenRouter & Provider routing, fallback configuration, returned fallback
model, and zero-data-retention routing constraints.
\href{https://openrouter.ai/docs/guides/routing/provider-selection}{10},
\href{https://openrouter.ai/docs/guides/features/zdr}{11},
\href{https://openrouter.ai/docs/guides/routing/model-fallbacks}{12} &
No surveyed tool-search trace or unified receipt. \\
vLLM & Operator-visible serving configuration and logs in self-hosted
deployments.
\href{https://docs.vllm.ai/en/latest/serving/openai_compatible_server.html}{37}
& No portable per-answer receipt by default. \\
SGLang & Operator-visible serving configuration and production metrics.
\href{https://docs.sglang.io/references/production_metrics.html}{38},
\href{https://arxiv.org/abs/2312.07104}{39} & No portable per-answer
receipt by default. \\
\end{longtable}

The table repurposes the vendor citations. They are not a literature
review. They are cell-level evidence for a small survey of
route-observability surfaces. The scholarly work below provides the
broader provenance, documentation, observability, and governance
context.

\begin{center}\rule{0.5\linewidth}{0.5pt}\end{center}

\hypertarget{why-model-names-are-not-enough}{%
\subsection{3. Why model names are not
enough}\label{why-model-names-are-not-enough}}

A model name helps, but it does not give you a complete audit trail.

Two requests can share the same product and visible model label while
hiding important differences: stable snapshot versus moving alias,
high-effort compute versus latency-optimized compute, searched answer
versus parameter-only answer, regional endpoint versus global endpoint,
intended provider versus fallback.

From the user's point of view, these may all look like ``the same AI.''
For accountability, they are distinct artifacts.

Version aliases make the difference explicit. A moving alias is
convenient because it lets users get improved models without changing
code. But it creates a reproducibility risk when exact reruns matter. A
benchmark run, a legal workflow, a medical summarization system, or a
production regression test cannot be reconstructed from a prompt and a
generic model label if the underlying version changed.

Service tiers create ambiguity. A user who pays for priority capacity is
buying both service and a model. If some requests are served as standard
because of ramp-rate behavior, overflow, quota exhaustion, or tier
assignment, the effective tier is part of what the user gets. When the
answer is correct and fast, this distinction is often ignored. In a
postmortem, it suddenly matters.

Tool use changes both how a response is produced and what evidence it
carries. A web-searched answer gives you sources to inspect and an
external dependency to question. A file-retrieval answer depends on the
retrieved material and the retrieval configuration. A code-execution
answer has an execution trace, or should. A no-tool answer may be more
stable and cheaper, but it can be less current. If the system decides
automatically whether to search, retrieve, or run code, that decision is
part of the route.

Fallback behavior can mitigate outages, route around rate limits, or
move a request to another model. OpenRouter's fallback documentation,
for example, says fallback can be triggered when primary model providers
are down, rate-limited, or refuse to reply because of content
moderation, and pricing is based on the model ultimately used, which is
returned in the response body.
(\href{https://openrouter.ai/docs/guides/routing/model-fallbacks}{OpenRouter})

A fallback can keep the system working, and it becomes part of the
record when someone later tries to reproduce, audit, or explain an
answer.

Saying ``the model answered'' leaves out how it was produced. In
adaptive systems, a service path produces it. That path may include
model selection, version resolution, service-tier assignment, effort
allocation, tool invocation, fallback, safety intervention,
data-location routing, and context management. Some of these choices are
irrelevant for a given request; others are decisive.

A route fact matters when it changes whether the answer can be trusted,
reproduced, governed, or challenged.

\begin{center}\rule{0.5\linewidth}{0.5pt}\end{center}

\hypertarget{routing-literature}{%
\subsection{4. Routing literature}\label{routing-literature}}

Routing literature explains how systems choose routes.

Language Model Cascades treated model calls, verifiers,
chain-of-thought-style steps, selection-inference, and tool use as
probabilistic programs. (\href{https://arxiv.org/abs/2207.10342}{arXiv})
FrugalGPT studied ways to reduce LLM inference cost, including cascades
that learn which combinations of models to use for different queries,
and found that cascaded strategies could match strong-model performance
with large cost reductions in some settings.
(\href{https://arxiv.org/abs/2305.05176}{arXiv}) RouteLLM learned
routers from preference data to choose between stronger and weaker
models, which cut cost while preserving response quality.
(\href{https://arxiv.org/abs/2406.18665}{arXiv})

This literature frames routing as a systems problem: choosing a model,
tool, or path that balances cost and quality. It also explains why many
product behaviors users now encounter exist.

A router can improve system performance while remaining opaque to users,
lowering average cost and improving reliability, but it can also make
similar prompts behave differently, hide fallback, and complicate
production debugging. Benchmark performance may hold while users still
lack enough agency to constrain routes in sensitive settings.

Recent benchmark work leaves routing quality unsettled. A 2026 preprint,
LLMRouterBench, presents a large benchmark and unified framework for LLM
routing. It confirms model complementarity but reports that many routing
methods perform similarly under unified evaluation and that several
recent approaches, including commercial routers, do not reliably
outperform a simple baseline.
(\href{https://arxiv.org/abs/2601.07206}{arXiv}) Routing deserves
observability because quality remains context-dependent and empirically
unsettled. Users and operators need records of what routers did.

Adaptive compute adds another layer to routing. A route is about more
than ``which model.'' It can also be ``how much computation did this
model receive?'' OpenAI's reasoning documentation describes reasoning
effort as a way to guide how much the model thinks, and model-specific
documentation includes effort settings such as low, medium, high, and
xhigh for some models.
(\href{https://developers.openai.com/api/docs/guides/reasoning}{OpenAI
API Documentation})
(\href{https://developers.openai.com/api/docs/models/gpt-5.3-codex}{OpenAI
API Documentation}) Vertex AI's generation configuration includes a
thinking budget that controls the tradeoff between response quality and
latency, and its routing configuration can select a model automatically
based on request content or manually by model name.
(\href{https://docs.cloud.google.com/vertex-ai/generative-ai/docs/reference/rest/v1/GenerationConfig}{Google
Cloud Vertex AI Documentation})

Two responses from the same named model can come from different compute
conditions, producing different costs, evaluation results, and safety
review needs. In a casual brainstorming session, that difference may not
matter much. It matters for cost accounting, evaluation, and
safety-critical review.

The transparency literature helps frame the problem. Model Cards
proposed short documents for trained models that cover intended use,
evaluation procedures, and performance characteristics.
(\href{https://arxiv.org/abs/1810.03993}{arXiv}) NIST's AI Risk
Management Framework frames AI trustworthiness as a system-level
risk-management problem.
(\href{https://www.nist.gov/itl/ai-risk-management-framework}{NIST})
OECD's AI Principles call for transparency and responsible disclosure,
including context-appropriate information.
(\href{https://www.oecd.org/en/topics/ai-principles.html}{OECD}) Work on
appropriate reliance says the goal is calibrated reliance: accepting
correct AI advice and rejecting incorrect advice.
(\href{https://arxiv.org/abs/2302.02187}{arXiv})

The missing artifact is a per-request record: model cards cover models,
governance frameworks cover responsibilities and risks, and routing
research covers optimization methods. None of them tells a developer
what route served a particular request at 14:03 on Tuesday.

The paper focuses on that missing per-request record.

\begin{center}\rule{0.5\linewidth}{0.5pt}\end{center}

\hypertarget{adapting-provenance-ideas-for-route-receipts}{%
\subsection{5. Adapting provenance ideas for route
receipts}\label{adapting-provenance-ideas-for-route-receipts}}

Route receipts adapt older provenance ideas for hosted AI inference. W3C
PROV-DM and PROV-O provide a general model for entities, activities,
agents, derivation, attribution, and generation.
(\href{https://www.w3.org/TR/prov-dm/}{W3C})
(\href{https://www.w3.org/TR/prov-o/}{W3C}) A route receipt is less
expressive than PROV, but it treats an answer as an artifact produced by
a runtime activity under particular conditions.

ML operations tooling uses the same approach. MLflow Tracking records
parameters, metrics, artifacts, code versions, and run metadata for
experiments.
(\href{https://mlflow.org/docs/latest/ml/tracking/}{MLflow}) SLSA
provenance records how software artifacts were built and under what
process guarantees. (\href{https://slsa.dev/spec/v1.2/provenance}{SLSA})
Route receipts bring that style of record to inference.

Route receipts belong with other documentation artifacts for AI systems.
Model Cards, Datasheets for Datasets, Data Statements for NLP, and IBM's
FactSheets make different parts of the AI lifecycle legible.
(\href{https://arxiv.org/abs/1810.03993}{arXiv})
(\href{https://arxiv.org/abs/1803.09010}{arXiv})
(\href{https://aclanthology.org/Q18-1041/}{ACL Anthology})
(\href{https://research.ibm.com/publications/factsheets-increasing-trust-in-ai-services-through-suppliers-declarations-of-conformity}{IBM
Research}) Those artifacts document models, datasets, language data, or
AI services. Route receipts document a single runtime event. A model
card says what a model is intended to do. A receipt says which route
served a particular request.

Production ML failures often come from system-level factors outside the
model itself. Hidden Technical Debt in Machine Learning Systems
identifies system-level risks such as data dependencies, configuration
debt, undeclared consumers, and changes in the external world.
(\href{https://papers.nips.cc/paper/5656-hidden-technical-debt-in-machine-learning-systems}{NeurIPS})
The ML Test Score makes monitoring and production-readiness explicit,
and TFX and data-validation work show how much production ML depends on
metadata, validation, and pipeline state.
(\href{https://research.google/pubs/the-ml-test-score-a-rubric-for-ml-production-readiness-and-technical-debt-reduction/}{Google
Research})
(\href{https://www.kdd.org/kdd2017/papers/view/tfx-a-tensorflow-based-production-scale-machine-learning-platform}{KDD})
(\href{https://proceedings.mlsys.org/paper_files/paper/2019/hash/928f1160e52192e3e0017fb63ab65391-Abstract.html}{MLSys})
Route receipts add that system-level view to LLM inference.

Industry observability tools point to the need for a portable per-answer
record. Fiddler, WhyLabs, and Arize Phoenix frame production AI
reliability around traces, monitoring, drift, evaluations, and
debugging. (\href{https://docs.fiddler.ai/observability}{Fiddler})
(\href{https://docs.whylabs.ai/docs/whylabs-overview-observe/}{WhyLabs})
(\href{https://arize.com/blog/llm-tracing-and-observability-with-arize-phoenix/}{Arize})
Route receipts add a minimum portable record that travels with an answer
even when the underlying observability stack is provider-specific.

OpenTelemetry's Generative AI semantic conventions overlap with this
proposal. They define telemetry attributes and spans for generative AI
calls, including requested and response model fields, provider names,
tool execution spans, and retrieval spans. Route receipts serve a
different layer from OpenTelemetry, which is mainly an instrumentation
and tracing standard for operators. A route receipt is a portable,
redacted, per-answer artifact that can be shown to users, exported with
benchmark records, or retained for audit without exposing the full
trace.
(\href{https://opentelemetry.io/docs/specs/semconv/gen-ai/}{OpenTelemetry})

The constraint also applies outside managed APIs. Open-source serving
stacks such as vLLM and SGLang expose OpenAI-compatible serving surfaces
and production metrics, but those metrics do not produce a portable
per-answer receipt.
(\href{https://docs.vllm.ai/en/latest/serving/openai_compatible_server.html}{vLLM})
(\href{https://docs.sglang.io/references/production_metrics.html}{SGLang})
(\href{https://arxiv.org/abs/2312.07104}{arXiv}) A receipt has to work
whether the route is controlled by a commercial platform, an enterprise
gateway, or a self-hosted inference stack.

The regulatory and standards context requires deployers to have enough
information to interpret AI-system outputs. EU AI Act Article 13
requires transparency and information for deployers of high-risk AI
systems.
(\href{https://eur-lex.europa.eu/legal-content/en/TXT/?uri=CELEX\%3A32024R1689}{EUR-Lex})
NIST's AI RMF treats transparency and accountability as system-level
characteristics, and ISO/IEC 42001 frames AI governance as a
management-system obligation.
(\href{https://www.nist.gov/itl/ai-risk-management-framework}{NIST})
(\href{https://www.iso.org/standard/42001}{ISO}) Route receipts record
facts that governance processes can use. They do not satisfy compliance
by themselves.

\begin{center}\rule{0.5\linewidth}{0.5pt}\end{center}

\hypertarget{what-a-route-receipt-records}{%
\subsection{6. What a route receipt
records}\label{what-a-route-receipt-records}}

A route receipt records the service path used to handle an AI request.

``Receipt'' is deliberate. It is a transaction record, not a supply
chain, pricing algorithm, or staff schedule. It gives the parties enough
detail to see what was purchased, when, under what terms, and at what
price.

A route receipt would serve a similar role for adaptive AI.

The proposal uses a minimal schema, and providers can add fields. These
fields distinguish requested route from effective route and make common
incidents investigable. The author maintains the reference specification
at \href{https://www.routereceipt.org/}{routereceipt.org} as a companion
specification for this paper. It is not an independent standard or
third-party endorsement. Appendix A contains the same proposed v0.1
schema, which lets the paper be reviewed without the external site.

The base receipt includes required fields for parser compatibility,
identity, time, route class, fallback status, safety status, region
class, completion status, and redactions:

\begin{itemize}
\tightlist
\item
  \texttt{schema\_version}
\item
  \texttt{receipt\_id}
\item
  \texttt{request\_id}
\item
  \texttt{served\_at}
\item
  \texttt{model\_identifier\_type}
\item
  \texttt{fallback}
\item
  \texttt{safety}
\item
  \texttt{region\_class}
\item
  \texttt{completion\_status}
\item
  \texttt{redactions}
\end{itemize}

Optional fields add detail when the product collects it:

\begin{itemize}
\tightlist
\item
  \texttt{requested\_model} and \texttt{resolved\_model} distinguish the
  visible model label from the served model or route.
\item
  \texttt{service\_tier} records requested and effective capacity class.
\item
  \texttt{effort} records requested effort and effective completion
  status without exposing hidden reasoning.
\item
  \texttt{tools} records tool classes, invocation counts, and redacted
  result references.
\item
  \texttt{context} records truncation and retrieval counts.
\item
  \texttt{provider\_chain} records provider hops for administrator or
  auditor views.
\item
  \texttt{retention\_class} records whether the event is ephemeral,
  ordinary, regulated, or under audit hold.
\end{itemize}

The schema records route facts, not explanations. It records whether a
moving alias resolved to a particular version, whether the requested
tier was used, and whether tools were used. Its non-disclosure boundary
is narrow: no hidden reasoning, no proprietary routing logic, no
bypassable safety detail, and no infrastructure topology.

Redaction rules belong on fields, not left implicit. A field can show
different details to end users, developers, administrators, and
auditors. Exact model versions, provider names, retrieval identifiers,
safety categories, or regional details can be generalized or redacted,
and the receipt should say that redaction occurred and why.

For a consumer product, the receipt should stay lightweight and easy to
skim. It might appear as expandable labels: ``web search used,''
``answered in fast mode,'' ``model updated since previous answer,''
``fallback used,'' or ``response restricted by safety policy.'' Most
users would ignore those labels. It should not turn everyday chat into a
compliance dashboard.

In an enterprise deployment, the receipt belongs in administrators' logs
and dashboards. Administrators want aggregate views of fallback
frequency, route-error correlations, global endpoint use, the model
versions that served regulated workflows, region-constraint violations,
and whether safety interventions increased after a model update.

Researchers use a receipt to check reproducibility and confirm what
actually ran. A benchmark result should record the prompt, model label,
temperature, output, whether the model label was fixed or moving,
whether fallbacks were allowed, whether tools were enabled, the effort
setting used, and whether the route changed during the run.

The receipt is intentionally incomplete. It follows that same
non-disclosure boundary and cannot make nondeterministic systems
perfectly reproducible. Its purpose is to record the serving conditions
a reasonable user needs to interpret the response.

The distinction matters because transparency arguments can easily go too
far. If transparency means ``publish everything,'' it will fail. Vendors
have legitimate reasons to protect safety systems, infrastructure
details, security-sensitive logic, and proprietary optimization methods.
If transparency means ``leave a usable record of consequential route
facts,'' the demand becomes more practical.

That framing makes the demand implementable without turning transparency
into disclosure of all internals.

\hypertarget{worked-mapping-example}{%
\subsubsection{6.1 Worked mapping
example}\label{worked-mapping-example}}

A receipt does not require providers to invent an entirely new
observability plane. It can be assembled from fields many systems
already expose. Normalize those fields into one per-answer record and
preserve redactions explicitly.

\begin{longtable}[]{@{}
  >{\raggedright\arraybackslash}p{(\columnwidth - 4\tabcolsep) * \real{0.3333}}
  >{\raggedright\arraybackslash}p{(\columnwidth - 4\tabcolsep) * \real{0.3333}}
  >{\raggedright\arraybackslash}p{(\columnwidth - 4\tabcolsep) * \real{0.3333}}@{}}
\toprule\noalign{}
\begin{minipage}[b]{\linewidth}\raggedright
Existing surface
\end{minipage} & \begin{minipage}[b]{\linewidth}\raggedright
Example exposed field
\end{minipage} & \begin{minipage}[b]{\linewidth}\raggedright
Receipt target
\end{minipage} \\
\midrule\noalign{}
\endhead
\bottomrule\noalign{}
\endlastfoot
OpenAI Priority processing & Effective service tier in the response
\href{https://developers.openai.com/api/docs/guides/priority-processing}{1}
& Effective service tier \\
Anthropic service tiers & Assigned tier in response usage
\href{https://platform.claude.com/docs/en/api/service-tiers}{2} &
Effective service tier \\
Amazon Bedrock service tiers & Resolved service tier in API, CloudTrail,
and metrics
\href{https://docs.aws.amazon.com/bedrock/latest/userguide/service-tiers-inference.html}{3}
& Effective service tier \\
OpenAI web search & Search-call output item and citations
\href{https://developers.openai.com/api/docs/guides/tools-web-search}{8}
& Tool-use summary \\
OpenRouter fallback & Returned model after fallback
\href{https://openrouter.ai/docs/guides/routing/model-fallbacks}{12} &
Resolved model and fallback target \\
\end{longtable}

For example, if a priority request is served through standard capacity
after a capacity fallback, the receipt fragment can record the requested
route and the effective route without exposing the router logic:

\begin{Shaded}
\begin{Highlighting}[]
\FunctionTok{\{}
  \DataTypeTok{"service\_tier"}\FunctionTok{:} \FunctionTok{\{}
    \DataTypeTok{"requested"}\FunctionTok{:} \StringTok{"priority"}\FunctionTok{,}
    \DataTypeTok{"effective"}\FunctionTok{:} \StringTok{"default"}\FunctionTok{,}
    \DataTypeTok{"change\_reason"}\FunctionTok{:} \StringTok{"capacity"}
  \FunctionTok{\},}
  \DataTypeTok{"fallback"}\FunctionTok{:} \FunctionTok{\{} \DataTypeTok{"status"}\FunctionTok{:} \StringTok{"occurred"}\FunctionTok{,} \DataTypeTok{"reason"}\FunctionTok{:} \StringTok{"capacity"} \FunctionTok{\},}
  \DataTypeTok{"redactions"}\FunctionTok{:} \OtherTok{[]}
\FunctionTok{\}}
\end{Highlighting}
\end{Shaded}

\begin{center}\rule{0.5\linewidth}{0.5pt}\end{center}

\hypertarget{case-study-how-the-legal-summary-changed}{%
\subsection{7. Case study: how the legal summary
changed}\label{case-study-how-the-legal-summary-changed}}

Northstar is a fictional legal-technology company that provides contract
summarization for enterprise customers. Its product ingests long
agreements, pulls out obligations and exceptions, and produces a memo
counsel can review. Users rely on it to save time, even though it does
not make final legal decisions.

During development, Northstar tests the system with a model alias called
\texttt{contract-pro-latest}. It performs well on an internal validation
set, so the team deploys the feature with priority processing to keep
latency low. File search is enabled so the model can retrieve clauses
from uploaded contracts. Web search is disabled so summaries stay
grounded in customer documents and internal policy material. Fallback is
enabled to keep the sales team available.

The system runs normally for two months, then the issue appears. The
memo treated the indemnity obligation as unconditional, even though the
carve-out was present in the contract.

Without a route receipt, the investigation is guesswork.

The prompt may have been wrong, the retrieval system may have missed the
clause, or the model version may have changed. The priority route may
have failed, and another model answered. Fallback may have used a model
with a smaller context window. An internal policy document may have been
retrieved instead of the relevant contract section. The model may have
hit an output budget and omitted the exception. The customer may have
uploaded a corrupted file. The issue may already have been present in
the original validation and gone unnoticed.

Each hypothesis draws on a different log source, and only some can be
tested. The visible model label \texttt{contract-pro-latest} may have
changed. The token count does not indicate which chunks were retrieved.
The response text does not disclose fallback or service tier. The
request timestamp does not show route state at that time.

With route receipts, the request can include the record:

\begin{Shaded}
\begin{Highlighting}[]
\FunctionTok{\{}
  \DataTypeTok{"requested\_model"}\FunctionTok{:} \StringTok{"contract{-}pro{-}latest"}\FunctionTok{,}
  \DataTypeTok{"resolved\_model"}\FunctionTok{:} \StringTok{"contract{-}pro{-}2026{-}04{-}18"}\FunctionTok{,}
  \DataTypeTok{"model\_identifier\_type"}\FunctionTok{:} \StringTok{"moving\_alias"}\FunctionTok{,}
  \DataTypeTok{"service\_tier"}\FunctionTok{:} \FunctionTok{\{}
    \DataTypeTok{"requested"}\FunctionTok{:} \StringTok{"priority"}\FunctionTok{,}
    \DataTypeTok{"effective"}\FunctionTok{:} \StringTok{"priority"}\FunctionTok{,}
    \DataTypeTok{"change\_reason"}\FunctionTok{:} \StringTok{"none"}
  \FunctionTok{\},}
  \DataTypeTok{"fallback"}\FunctionTok{:} \FunctionTok{\{}
    \DataTypeTok{"status"}\FunctionTok{:} \StringTok{"none"}\FunctionTok{,}
    \DataTypeTok{"reason"}\FunctionTok{:} \StringTok{"none"}
  \FunctionTok{\},}
  \DataTypeTok{"tools"}\FunctionTok{:} \FunctionTok{\{}
    \DataTypeTok{"used"}\FunctionTok{:} \OtherTok{[}
      \FunctionTok{\{}
        \DataTypeTok{"name"}\FunctionTok{:} \StringTok{"file\_search"}\FunctionTok{,}
        \DataTypeTok{"invocation\_count"}\FunctionTok{:} \DecValTok{1}\FunctionTok{,}
        \DataTypeTok{"result\_refs"}\FunctionTok{:} \OtherTok{[}
          \StringTok{"contract\_chunks[14]"}\OtherTok{,}
          \StringTok{"contract\_chunks[15]"}\OtherTok{,}
          \StringTok{"contract\_chunks[22]"}\OtherTok{,}
          \StringTok{"contract\_chunks[41]"}\OtherTok{,}
          \StringTok{"policy\_chunks[3]"}\OtherTok{,}
          \StringTok{"policy\_chunks[8]"}
        \OtherTok{]}
      \FunctionTok{\}}
    \OtherTok{]}
  \FunctionTok{\},}
  \DataTypeTok{"context"}\FunctionTok{:} \FunctionTok{\{}
    \DataTypeTok{"input\_truncated"}\FunctionTok{:} \StringTok{"false"}\FunctionTok{,}
    \DataTypeTok{"retrieved\_item\_count"}\FunctionTok{:} \DecValTok{6}
  \FunctionTok{\},}
  \DataTypeTok{"effort"}\FunctionTok{:} \FunctionTok{\{}
    \DataTypeTok{"requested"}\FunctionTok{:} \StringTok{"high"}\FunctionTok{,}
    \DataTypeTok{"effective\_status"}\FunctionTok{:} \StringTok{"completed"}
  \FunctionTok{\},}
  \DataTypeTok{"completion\_status"}\FunctionTok{:} \StringTok{"complete"}\FunctionTok{,}
  \DataTypeTok{"safety"}\FunctionTok{:} \FunctionTok{\{}
    \DataTypeTok{"status"}\FunctionTok{:} \StringTok{"none"}\FunctionTok{,}
    \DataTypeTok{"visible\_action"}\FunctionTok{:} \StringTok{"none"}
  \FunctionTok{\},}
  \DataTypeTok{"region\_class"}\FunctionTok{:} \StringTok{"user\_selected\_region"}
\FunctionTok{\}}
\end{Highlighting}
\end{Shaded}

That receipt narrows the problem. The team can stop investigating
fallback and rule out a tier downgrade. The remaining checks are
narrower: inspect chunk 41 for the carve-out, compare the served model
with the validation snapshot, and rerun against the pinned prior
snapshot with the same retrieval set.

Validation ran against \texttt{contract-pro-2026-03-02}, while the
production request resolved to \texttt{contract-pro-2026-04-18}, so
production used the newer model, which summarizes exceptions more
aggressively. The fix is to pin a model version for this workflow, add a
regression test for indemnity carve-outs, or require a separate
extraction pass for exceptions. The receipt shows what happened, not
what the team should do.

The fictional case is controlled, but public incidents show model
updates can materially change production behavior. In April 2025, OpenAI
rolled back a GPT-4o update after users reported a noticeable shift
toward sycophancy; OpenAI later described the update, the rollback, and
changes it planned for evaluation and release processes.
(\href{https://openai.com/index/sycophancy-in-gpt-4o/}{OpenAI}) That
incident was not a route-receipt failure. It shows why production users
need records that distinguish fixed snapshots, moving aliases, and
updated runtime behavior when an answer matters.

A second Northstar incident shows the same issue. A customer reports
that some summaries are slow and inconsistent. Route receipts show that
most requests use priority capacity, and during traffic spikes some
shift to standard capacity. Another cluster shows fallback to a smaller
model when the preferred model is unavailable. The company disables
fallback for regulated customers, keeps it for low-stakes workflows, and
warns when it does.

Stable routes are not always the right choice. Northstar may still want
adaptive routing for draft emails, customer support, and internal
knowledge search. Different workflows need different route promises. A
legal summarization tool does not need maximum adaptiveness all the
time.

User agency means enforceable route constraints and records that show
whether they held.

\begin{center}\rule{0.5\linewidth}{0.5pt}\end{center}

\hypertarget{what-a-receipt-should-cover}{%
\subsection{8. What a receipt should
cover}\label{what-a-receipt-should-cover}}

A route receipt should reflect why someone might need it.

A developer debugging a request needs to know whether the requested
model was the served model, whether the model identifier was fixed or
moving, whether the service tier changed, whether fallback occurred,
whether tools were used, whether the context was truncated, whether the
response was incomplete, or whether a safety intervention affected the
result. The receipt should surface those conditions. A professional user
may need only a few visible indicators. An administrator may need
aggregate logs and policy controls. A researcher needs enough detail to
make evaluation runs comparable.

A receipt should stay clear of a few traps.

A receipt should not expose chain-of-thought. It should record that high
effort was requested, that the response completed, or that the budget
was exhausted. It should not reveal private reasoning traces, even when
they influenced the answer.

The same non-disclosure boundary applies to router behavior. Users
usually do not need scoring formulas, provider rankings, safety
thresholds, or traffic-shaping rules. A receipt can say ``fallback
occurred because primary provider was rate-limited'' without publishing
the threshold. It can say ``served through global endpoint'' without
naming the processing region when the provider's product does not
guarantee that region.

Receipts should stay honest about reproducibility. Even a detailed
receipt cannot guarantee that the same answer can be regenerated.
Sampling, external search results, file indexes, provider changes,
caching, rate limits, and nondeterministic serving behavior can all
affect outputs. A receipt records the material serving conditions.

Partial reconstruction can move a team beyond guesswork. In many
incidents, the goal is to identify which class of failure occurred,
whether that was a model-version regression, a retrieval miss, a
fallback effect, a budget exhaustion, a safety intervention, or a
region-policy violation. A receipt turns a vague complaint into an event
the team can investigate.

The receipt should record what was absent so the team can see which
route components were active and which were missing. ``No fallback'' is
information. ``No web search'' is information. ``No safety
intervention'' is information. ``Context not truncated'' is information.
Users often need to know which route components were active and which
were absent.

\begin{center}\rule{0.5\linewidth}{0.5pt}\end{center}

\hypertarget{route-transparency-and-user-trust}{%
\subsection{9. Route transparency and user
trust}\label{route-transparency-and-user-trust}}

Route transparency helps users trust AI systems more appropriately.

Research on appropriate reliance shows that people using AI advice need
to accept correct advice and reject incorrect advice. Explanations and
interface cues keep reliance calibrated or let it drift out of
alignment. (\href{https://arxiv.org/abs/2302.02187}{arXiv}) Route
receipts extend the same idea to adaptive service paths. Users should
not trust the answer just because ``the AI answered.'' Route receipts
make the route visible so users can judge the answer with the right
level of confidence. Users should understand enough about the route to
decide how much weight to give the answer.

A searched answer calls for a different review than an unsearched
answer. A fallback answer may warrant less confidence than an answer
served by the intended route. A response from a moving alias may work
for brainstorming, but it can fail a locked evaluation. A global
endpoint may suit ordinary consumer use but fail a workload with strict
processing-location requirements.

Receipts should fit the audience. In consumer chat, too much metadata
becomes noise, so a small indicator that search was used or a file was
consulted is often enough. In professional products, users may need
visible mode and route cues because cost and quality expectations matter
to the work. In developer APIs, structured metadata should be standard.
In enterprise settings, route facts belong in logs as well as UI badges.

The interface should stay usable for people who are not infrastructure
engineers. If the route materially changes the answer, surface it.

\begin{center}\rule{0.5\linewidth}{0.5pt}\end{center}

\hypertarget{usability-and-tradeoffs}{%
\subsection{10. Usability and tradeoffs}\label{usability-and-tradeoffs}}

Users often resist route receipts when the metadata feels unnecessary.

Many users feel that way on routine tasks. People asking for a recipe, a
joke, or a first draft of an email do not want to inspect service-tier
metadata. Even sophisticated users can find too many labels distracting.
If a route receipt is too prominent, the product can feel anxious,
bureaucratic, or broken.

The interface should present information by audience and task.

At the surface, the product can show only the route facts that change
the answer: web used, files consulted, fallback occurred, model changed,
response incomplete, safety restriction applied. The developer view can
expose structured route metadata for the same request. The admin view
can show route facts as aggregate trends and audit events. In research
mode, route facts can be exported with outputs.

Receipts should stay unobtrusive by default and become detailed only
when someone needs the record.

Receipts must limit what they reveal because the next objection is
sensitive information. Safety systems should not reveal bypassable
thresholds. Routers should not publish proprietary scoring functions.
Providers should not disclose security-sensitive failover behavior. Some
route facts may need to be generalized to protect sensitive details. A
receipt can say ``capacity fallback'' instead of naming the failed
provider. It might say ``safety category: sensitive-data masking''
instead of explaining how to avoid the mask.

Compliance theater weakens receipts. A vendor could produce lots of
metadata and still make poor routing choices. Documentation does not
replace governance, and a receipt cannot make a bad route good. It does
change who can see and challenge the routing decision. Without a
receipt, users may not know which route to challenge.

The engineering burden is real. Route receipts require schemas, logging,
access control, retention policy, UI design, and support workflows. Yet
many fragments already exist in service-tier fields, model versions,
tool traces, citation annotations, provider preferences, region
settings, fallback responses, and audit logs. The proposal asks
platforms to treat these fragments as one coherent provenance layer.

Some users may care only about outputs. If the answer is good, why does
the route matter?

In serious settings, route conditions determine whether a correct answer
is acceptable. A correct answer served outside an agreed data zone may
violate a customer requirement. A good answer produced by an unapproved
provider may break an enterprise policy. A high-quality benchmark answer
from a moving alias may be unusable for reproducibility. A fast answer
served through an unexpected tier may create billing or SLA questions.

\begin{center}\rule{0.5\linewidth}{0.5pt}\end{center}

\hypertarget{threat-model-and-redaction}{%
\subsection{11. Threat model and
redaction}\label{threat-model-and-redaction}}

A receipt works when it gives each audience the facts it needs. The same
field can be harmless in an internal audit log and harmful in a public
UI. Four stakeholder views define the minimal design: end user,
developer, administrator, and auditor. Four adversary pressures are
enough to test the design: curiosity, gaming, competition, and
regulatory review.

End users should see compact labels like model changed, search used,
fallback occurred, or safety restriction applied, not safety thresholds,
router scores, prompt filters, provider chains, capacity internals, or
infrastructure names unless the product explicitly promises those
details.

Developers should see request IDs, the resolved model or model class,
tool classes, completion status, retrieval counts, and coarse safety or
fallback reasons. Ordinary developer logs should still avoid bypassable
policy details, provider contracts, pricing rules, and exact routing
weights.

Administrators should see aggregate fallback rates, region classes, tier
changes, policy hits, retention classes, and suspicious probing
patterns. Their exports should use contractual labels and coarse route
classes unless a stronger access-control regime is in place.

Auditors should receive the most complete view, including the full
receipt where allowed, redaction rationale, policy mapping,
access-control evidence, and fields omitted from user or developer
views. Safety-sensitive fields can still be reviewed in controlled
environments rather than exposed in the product UI.

This matrix implies three disclosure tiers. The public tier contains
labels that help a user interpret an answer. The operational tier
contains route facts needed for debugging and administration. The audit
tier contains the most complete record, including redaction reasons and
policy mappings. The same receipt object can support all three by
marking field-level redactions instead of pretending that one view fits
every case.

\texttt{redactions} should be a required schema field in the receipt. If
the resolved provider, exact safety category, retrieval identifiers, or
region detail is hidden, the receipt should mark those fields redacted
and explain why. Silence leaves the record ambiguous. A redaction notice
makes the absence explicit and accountable in the receipt.

\begin{center}\rule{0.5\linewidth}{0.5pt}\end{center}

\hypertarget{route-constraints-as-a-product-promise}{%
\subsection{12. Route constraints as a product
promise}\label{route-constraints-as-a-product-promise}}

``Powered by Model X'' will not disappear. It is too useful as a product
label. In routed systems, the label often omits the route constraints. A
clearer promise is:

Choose the route constraints that matter for your use case. You can
optimize for cheap drafts, stable evaluations, provider limits, tool
access, or data-location controls. After the system serves your request,
you can see enough about the effective route to verify whether those
preferences held. When they did not hold, the system records the
difference in a way that can be debugged.

Several providers already show parts of this promise. OpenAI records
which service tier processed priority requests and how it downgrades
under ramp-rate limits. Anthropic's response usage object includes the
assigned service tier. AWS Bedrock exposes the service tier through
response and monitoring surfaces, and its prompt-routing feature returns
information about the model used. Google distinguishes stable and latest
model identifiers, and warns that global endpoints do not guarantee
control over processing region. OpenRouter lets users constrain provider
routing and zero-data-retention settings.
(\href{https://developers.openai.com/api/docs/guides/priority-processing}{OpenAI
API Documentation})
(\href{https://platform.claude.com/docs/en/api/service-tiers}{Anthropic
API Documentation})
(\href{https://docs.aws.amazon.com/bedrock/latest/userguide/service-tiers-inference.html}{AWS
Documentation})
(\href{https://docs.aws.amazon.com/bedrock/latest/userguide/prompt-routing.html}{AWS
Documentation})
(\href{https://ai.google.dev/gemini-api/docs/models}{Google AI for
Developers})
(\href{https://docs.cloud.google.com/vertex-ai/generative-ai/docs/learn/locations}{Google
Cloud Vertex AI Documentation})
(\href{https://openrouter.ai/docs/guides/routing/provider-selection}{OpenRouter})
(\href{https://openrouter.ai/docs/guides/features/zdr}{OpenRouter})

These features come from different products and serve different
purposes, but they show that adaptive AI systems can expose route facts
without exposing router internals.

The proposal here is to define those fragments as one small record.

\begin{center}\rule{0.5\linewidth}{0.5pt}\end{center}

\hypertarget{route-transparency-and-model-transparency}{%
\subsection{13. Route transparency and model
transparency}\label{route-transparency-and-model-transparency}}

Model cards remain useful. So do system cards, evaluations, incident
reports, lifecycle notices, and risk-management frameworks. A route
receipt does not replace any of them.

The receipt covers runtime facts those documents usually do not cover.

A model card describes a trained artifact: intended uses, evaluation
results, limitations, and performance across conditions. A lifecycle
notice covers support status and service changes. A service-tier guide
covers capacity classes and limits. A tool-use guide covers tool
capabilities and constraints. A data-location document covers endpoint
behavior and routing.

A route receipt says something about a particular answer.

AI accountability depends on design-time and runtime information from
both documentation and receipts. Design-time documentation states what a
system is intended to do. Runtime receipts show what the system did. In
routed AI, the questions differ.

Route receipts record which service path handled each request for
governance teams. NIST's AI RMF calls for managing AI risks across the
lifecycle and treats transparency and accountability as system-level
characteristics.
(\href{https://www.nist.gov/itl/ai-risk-management-framework}{NIST})
OECD's principles call for meaningful information appropriate to
context.
(\href{https://www.oecd.org/en/topics/ai-principles.html}{OECD}) In
hosted AI systems, route receipts make the runtime service path visible
for challenge, debugging, and accountability.

``Appropriate to context'' means the receipt should fit the setting and
audience. A route receipt for a consumer assistant, a developer API, a
public-sector procurement system, and a benchmark harness should show
different detail levels while preserving the event record.

\begin{center}\rule{0.5\linewidth}{0.5pt}\end{center}

\hypertarget{limitations-and-future-work}{%
\subsection{14. Limitations and future
work}\label{limitations-and-future-work}}

This paper is a position paper with a schema proposal and a
documentation-based survey. It does not yet contain live measurements of
alias drift, fallback frequency, regional routing, tool-use routing, or
compute-budget effects. The survey table should therefore be read as
evidence of current observability surfaces, not as a claim about how
often those routes occur in production.

The strongest next empirical contribution would be narrow and
repeatable. One option is an alias-drift probe: query a small set of
fixed prompts against platforms that publish moving aliases, log the
resolved model where exposed, and report drift events over a defined
period. A second option is a fallback-observability probe: deliberately
trigger rate limits, provider errors, malformed inputs, or unavailable
models where allowed by terms, then measure whether the effective
fallback is visible from the API response, logs, or dashboard. Either
study would turn the survey in Table 1 into a baseline for measurement.

Several design issues also need deeper analysis. Consent should be
modeled by workflow: ordinary users do not need to approve every
capacity decision, while regulated customers may need to prohibit
provider changes, global routing, tool use, or fallback to unapproved
models. Adaptive compute needs vocabulary that records requested effort,
effective status, and budget exhaustion without exposing hidden
reasoning traces. Safety interventions need category-level disclosure
that supports accountability without teaching attackers how to probe
guardrails. Moving aliases need lifecycle notices and regression-testing
hooks, not only after-the-fact receipts. Retention policy must balance
auditability with privacy because receipts can contain sensitive
operational and user metadata.

The Northstar example is illustrative rather than an incident report.
Future work should replace or supplement it with a small corpus of
public incidents involving model updates, route changes, fallback
behavior, or version confusion. The purpose would not be to assign blame
to providers. It would be to test whether route receipts would have made
each incident easier to detect, explain, or reproduce.

\begin{center}\rule{0.5\linewidth}{0.5pt}\end{center}

\hypertarget{conclusion}{%
\subsection{15. Conclusion}\label{conclusion}}

Adaptive AI services need runtime transparency alongside model-level
documentation. This paper introduced the route receipt as a compact
record of the service path that handled a particular answer. The
proposal contributes a minimal schema, a redaction model, and a
documentation-based survey showing that selected platforms already
expose fragments of this record but not a portable receipt.

The paper deliberately does less than a benchmark or field study. It
does not measure alias drift, fallback frequency, regional routing,
tool-use routing, or compute-budget effects in live systems. It also
does not argue for disclosure of proprietary router logic or hidden
reasoning. The narrower claim is that routed AI systems should leave a
usable event record for consequential route facts.

The next step is empirical. A small alias-drift probe, a
fallback-observability probe, or a corpus of public incidents would test
whether the schema captures the facts developers and auditors actually
need. Those additions would turn the route receipt from a design
proposal into an empirically grounded transparency artifact for routed
AI systems.

\begin{center}\rule{0.5\linewidth}{0.5pt}\end{center}

\hypertarget{references}{%
\subsection{References}\label{references}}

\begin{enumerate}
\def\labelenumi{\arabic{enumi}.}
\tightlist
\item
  OpenAI API Documentation.
  \href{https://developers.openai.com/api/docs/guides/priority-processing}{``Priority
  processing \textbar{} OpenAI API''}. Accessed April 29, 2026.
\item
  Anthropic API Documentation.
  \href{https://platform.claude.com/docs/en/api/service-tiers}{``Service
  tiers''}. Accessed April 29, 2026.
\item
  AWS Documentation.
  \href{https://docs.aws.amazon.com/bedrock/latest/userguide/service-tiers-inference.html}{``Service
  tiers for optimizing performance and cost''}. Accessed April 29, 2026.
\item
  Google AI for Developers.
  \href{https://ai.google.dev/gemini-api/docs/models}{``Models
  \textbar{} Gemini API''}. Accessed April 29, 2026.
\item
  Microsoft Learn.
  \href{https://learn.microsoft.com/en-us/azure/foundry/openai/concepts/model-retirements}{``Foundry
  Models lifecycle and support policy''}. Accessed April 29, 2026.
\item
  AWS Documentation.
  \href{https://docs.aws.amazon.com/bedrock/latest/userguide/prompt-routing.html}{``Understanding
  intelligent prompt routing in Amazon Bedrock''}. Accessed April 29,
  2026.
\item
  Microsoft Learn.
  \href{https://learn.microsoft.com/en-us/azure/foundry/openai/concepts/model-router}{``Model
  router for Microsoft Foundry concepts''}. Accessed April 29, 2026.
\item
  OpenAI API Documentation.
  \href{https://developers.openai.com/api/docs/guides/tools-web-search}{``Web
  search \textbar{} OpenAI API''}. Accessed April 29, 2026.
\item
  Google Cloud Vertex AI Documentation.
  \href{https://docs.cloud.google.com/vertex-ai/generative-ai/docs/learn/locations}{``Deployments
  and endpoints \textbar{} Generative AI on Vertex AI''}. Accessed April
  29, 2026.
\item
  OpenRouter Documentation.
  \href{https://openrouter.ai/docs/guides/routing/provider-selection}{``Provider
  Routing''}. Accessed April 29, 2026.
\item
  OpenRouter Documentation.
  \href{https://openrouter.ai/docs/guides/features/zdr}{``Zero Data
  Retention''}. Accessed April 29, 2026.
\item
  OpenRouter Documentation.
  \href{https://openrouter.ai/docs/guides/routing/model-fallbacks}{``Model
  Fallbacks''}. Accessed April 29, 2026.
\item
  David Dohan, Winnie Xu, Aitor Lewkowycz, Jacob Austin, David Bieber,
  Raphael Gontijo Lopes, Yuhuai Wu, Henryk Michalewski, Rif A. Saurous,
  Jascha Sohl-Dickstein, Kevin Murphy, and Charles Sutton.
  \href{https://arxiv.org/abs/2207.10342}{``Language Model Cascades''}.
  arXiv:2207.10342, 2022.
\item
  Lingjiao Chen, Matei Zaharia, and James Zou.
  \href{https://arxiv.org/abs/2305.05176}{``FrugalGPT: How to Use Large
  Language Models While Reducing Cost and Improving Performance''}.
  arXiv:2305.05176, 2023.
\item
  Isaac Ong, Amjad Almahairi, Vincent Wu, Wei-Lin Chiang, Tianhao Wu,
  Joseph E. Gonzalez, M. Waleed Kadous, and Ion Stoica.
  \href{https://arxiv.org/abs/2406.18665}{``RouteLLM: Learning to Route
  LLMs with Preference Data''}. arXiv:2406.18665, 2024.
\item
  Hao Li, Yiqun Zhang, Zhaoyan Guo, Chenxu Wang, Shengji Tang, Qiaosheng
  Zhang, Yang Chen, Biqing Qi, Peng Ye, Lei Bai, Zhen Wang, and Shuyue
  Hu. \href{https://arxiv.org/abs/2601.07206}{``LLMRouterBench: A
  Massive Benchmark and Unified Framework for LLM Routing''}.
  arXiv:2601.07206, 2026.
\item
  OpenAI API Documentation.
  \href{https://developers.openai.com/api/docs/guides/reasoning}{``Reasoning
  models \textbar{} OpenAI API''}. Accessed April 29, 2026.
\item
  Google Cloud Vertex AI Documentation.
  \href{https://docs.cloud.google.com/vertex-ai/generative-ai/docs/reference/rest/v1/GenerationConfig}{``GenerationConfig
  \textbar{} Generative AI on Vertex AI''}. Accessed April 29, 2026.
\item
  Margaret Mitchell, Simone Wu, Andrew Zaldivar, Parker Barnes, Lucy
  Vasserman, Ben Hutchinson, Elena Spitzer, Inioluwa Deborah Raji, and
  Timnit Gebru. \href{https://arxiv.org/abs/1810.03993}{``Model Cards
  for Model Reporting''}. Proceedings of the Conference on Fairness,
  Accountability, and Transparency (FAT*), 2019.
\item
  National Institute of Standards and Technology.
  \href{https://www.nist.gov/itl/ai-risk-management-framework}{``Artificial
  Intelligence Risk Management Framework (AI RMF 1.0)''}. NIST AI 100-1,
  2023.
\item
  OECD. \href{https://www.oecd.org/en/topics/ai-principles.html}{``OECD
  AI Principles''}. OECD Recommendation on Artificial Intelligence,
  2019, updated 2024.
\item
  Max Schemmer, Niklas Kühl, Carina Benz, Andrea Bartos, and Gerhard
  Satzger. \href{https://arxiv.org/abs/2302.02187}{``Appropriate
  Reliance on AI Advice: Conceptualization and the Effect of
  Explanations''}. Proceedings of the 28th International Conference on
  Intelligent User Interfaces (IUI), 2023.
\item
  W3C. \href{https://www.w3.org/TR/prov-dm/}{``PROV-DM: The PROV Data
  Model''}. W3C Recommendation, 2013.
\item
  W3C. \href{https://www.w3.org/TR/prov-o/}{``PROV-O: The PROV
  Ontology''}. W3C Recommendation, 2013.
\item
  MLflow Documentation.
  \href{https://mlflow.org/docs/latest/ml/tracking/}{``MLflow
  Tracking''}. Accessed April 29, 2026.
\item
  SLSA. \href{https://slsa.dev/spec/v1.2/provenance}{``SLSA
  Provenance''}. Version 1.2. Accessed April 29, 2026.
\item
  Timnit Gebru, Jamie Morgenstern, Briana Vecchione, Jennifer Wortman
  Vaughan, Hanna Wallach, Hal Daume III, and Kate Crawford.
  \href{https://arxiv.org/abs/1803.09010}{``Datasheets for Datasets''}.
  Communications of the ACM, 64(12), 2021.
\item
  Emily M. Bender and Batya Friedman.
  \href{https://aclanthology.org/Q18-1041/}{``Data Statements for
  Natural Language Processing''}. Transactions of the Association for
  Computational Linguistics, 6, 2018.
\item
  Matthew Arnold, Rachel K. E. Bellamy, Michael Hind, Stephanie Houde,
  Sameep Mehta, Aleksandra Mojsilović, Ravi Nair, Karthikeyan Natesan
  Ramamurthy, Darrell Reimer, Alexandra Olteanu, David Piorkowski, Jason
  Tsay, and Kush R. Varshney.
  \href{https://research.ibm.com/publications/factsheets-increasing-trust-in-ai-services-through-suppliers-declarations-of-conformity}{``FactSheets:
  Increasing trust in AI services through supplier's declarations of
  conformity''}. IBM Journal of Research and Development, 63(4/5), 2019.
\item
  D. Sculley, Gary Holt, Daniel Golovin, Eugene Davydov, Todd Phillips,
  Dietmar Ebner, Vinay Chaudhary, Michael Young, Jean-Francois Crespo,
  and Dan Dennison.
  \href{https://papers.nips.cc/paper/5656-hidden-technical-debt-in-machine-learning-systems}{``Hidden
  Technical Debt in Machine Learning Systems''}. Advances in Neural
  Information Processing Systems 28 (NIPS), 2015.
\item
  Eric Breck, Shanqing Cai, Eric Nielsen, Michael Salib, and D. Sculley.
  \href{https://research.google/pubs/the-ml-test-score-a-rubric-for-ml-production-readiness-and-technical-debt-reduction/}{``The
  ML Test Score: A Rubric for ML Production Readiness and Technical Debt
  Reduction''}. IEEE International Conference on Big Data, 2017.
\item
  Denis Baylor, Eric Breck, Heng-Tze Cheng, Noah Fiedel, Chuan Yu Foo,
  Zakaria Haque, Salem Haykal, Mustafa Ispir, Vihan Jain, Levent Koc,
  Chiu Yuen Koo, Lukasz Lew, Clemens Mewald, Akshay Naresh Modi, Neoklis
  Polyzotis, Sukriti Ramesh, Sudip Roy, Steven Euijong Whang, Martin
  Wicke, Jarek Wilkiewicz, Xin Zhang, and Martin Zinkevich.
  \href{https://www.kdd.org/kdd2017/papers/view/tfx-a-tensorflow-based-production-scale-machine-learning-platform}{``TFX:
  A TensorFlow-Based Production-Scale Machine Learning Platform''}.
  Proceedings of KDD, 2017.
\item
  Eric Breck, Martin Zinkevich, Neoklis Polyzotis, Steven Whang, and
  Sudip Roy.
  \href{https://proceedings.mlsys.org/paper_files/paper/2019/hash/928f1160e52192e3e0017fb63ab65391-Abstract.html}{``Data
  Validation for Machine Learning''}. Proceedings of MLSys, 2019.
\item
  Fiddler Documentation.
  \href{https://docs.fiddler.ai/observability}{``Fiddler
  Observability''}.
\item
  WhyLabs Documentation.
  \href{https://docs.whylabs.ai/docs/whylabs-overview-observe/}{``WhyLabs
  Observe''}.
\item
  Arize.
  \href{https://arize.com/blog/llm-tracing-and-observability-with-arize-phoenix/}{``LLM
  Tracing and Observability with Arize Phoenix''}.
\item
  vLLM Documentation.
  \href{https://docs.vllm.ai/en/latest/serving/openai_compatible_server.html}{``OpenAI-Compatible
  Server''}. Accessed April 29, 2026.
\item
  SGLang Documentation.
  \href{https://docs.sglang.io/references/production_metrics.html}{``Production
  Metrics''}. Accessed April 29, 2026.
\item
  Zheng et al.~\href{https://arxiv.org/abs/2312.07104}{``SGLang:
  Efficient Execution of Structured Language Model Programs''}.
\item
  European Union.
  \href{https://eur-lex.europa.eu/legal-content/en/TXT/?uri=CELEX\%3A32024R1689}{``Regulation
  (EU) 2024/1689''}.
\item
  ISO. \href{https://www.iso.org/standard/42001}{``ISO/IEC
  42001:2023''}.
\item
  OpenAI.
  \href{https://openai.com/index/sycophancy-in-gpt-4o/}{``Sycophancy in
  GPT-4o: what happened and what we're doing about it''}.
\item
  OpenAI API Documentation.
  \href{https://developers.openai.com/api/docs/guides/your-data}{``Data
  controls in the OpenAI platform''}. Accessed April 29, 2026.
\item
  Anthropic API Documentation.
  \href{https://platform.claude.com/docs/en/agents-and-tools/tool-use/web-search-tool}{``Web
  search tool''}. Accessed April 29, 2026.
\item
  OpenAI API Documentation.
  \href{https://developers.openai.com/api/docs/models/gpt-5.3-codex}{``GPT-5.3-Codex
  Model \textbar{} OpenAI API''}. Accessed April 29, 2026.
\item
  Route Receipt Specification.
  \href{https://www.routereceipt.org/}{``Route Receipt Specification''}.
  Maintained by Vincent Schmalbach. Accessed April 30, 2026.
\item
  OpenTelemetry.
  \href{https://opentelemetry.io/docs/specs/semconv/gen-ai/}{``Semantic
  conventions for generative AI systems''}. Accessed May 1, 2026.
\end{enumerate}

\begin{center}\rule{0.5\linewidth}{0.5pt}\end{center}

\hypertarget{appendix-a-minimal-route-receipt-json-schema}{%
\subsection{Appendix A: Minimal route receipt JSON
Schema}\label{appendix-a-minimal-route-receipt-json-schema}}

This schema is intentionally small. It defines a portable receipt object
that can be embedded in API responses, logs, audit exports, or benchmark
records. Providers can add extension fields under
\texttt{provider\_extensions}, but the base fields should keep the same
meaning across systems. The canonical v0.1 schema URL is
\texttt{https://routereceipt.org/schemas/route-receipt/v0.1/schema.json}.
Subsequent schema versions should follow semantic versioning, with
additions kept backward-compatible within a major version.

\begin{Shaded}
\begin{Highlighting}[]
\FunctionTok{\{}
  \DataTypeTok{"$schema"}\FunctionTok{:} \StringTok{"https://json{-}schema.org/draft/2020{-}12/schema"}\FunctionTok{,}
  \DataTypeTok{"$id"}\FunctionTok{:} \StringTok{"https://routereceipt.org/schemas/route{-}receipt/v0.1/schema.json"}\FunctionTok{,}
  \DataTypeTok{"title"}\FunctionTok{:} \StringTok{"RouteReceipt"}\FunctionTok{,}
  \DataTypeTok{"type"}\FunctionTok{:} \StringTok{"object"}\FunctionTok{,}
  \DataTypeTok{"additionalProperties"}\FunctionTok{:} \KeywordTok{false}\FunctionTok{,}
  \DataTypeTok{"required"}\FunctionTok{:} \OtherTok{[}
    \StringTok{"schema\_version"}\OtherTok{,}
    \StringTok{"receipt\_id"}\OtherTok{,}
    \StringTok{"request\_id"}\OtherTok{,}
    \StringTok{"served\_at"}\OtherTok{,}
    \StringTok{"model\_identifier\_type"}\OtherTok{,}
    \StringTok{"fallback"}\OtherTok{,}
    \StringTok{"safety"}\OtherTok{,}
    \StringTok{"region\_class"}\OtherTok{,}
    \StringTok{"completion\_status"}\OtherTok{,}
    \StringTok{"redactions"}
  \OtherTok{]}\FunctionTok{,}
  \DataTypeTok{"properties"}\FunctionTok{:} \FunctionTok{\{}
    \DataTypeTok{"schema\_version"}\FunctionTok{:} \FunctionTok{\{}
      \DataTypeTok{"type"}\FunctionTok{:} \StringTok{"string"}\FunctionTok{,}
      \DataTypeTok{"const"}\FunctionTok{:} \StringTok{"route{-}receipt.v0.1"}
    \FunctionTok{\},}
    \DataTypeTok{"receipt\_id"}\FunctionTok{:} \FunctionTok{\{}
      \DataTypeTok{"type"}\FunctionTok{:} \StringTok{"string"}\FunctionTok{,}
      \DataTypeTok{"minLength"}\FunctionTok{:} \DecValTok{1}
    \FunctionTok{\},}
    \DataTypeTok{"request\_id"}\FunctionTok{:} \FunctionTok{\{}
      \DataTypeTok{"type"}\FunctionTok{:} \StringTok{"string"}\FunctionTok{,}
      \DataTypeTok{"minLength"}\FunctionTok{:} \DecValTok{1}
    \FunctionTok{\},}
    \DataTypeTok{"served\_at"}\FunctionTok{:} \FunctionTok{\{}
      \DataTypeTok{"type"}\FunctionTok{:} \StringTok{"string"}\FunctionTok{,}
      \DataTypeTok{"format"}\FunctionTok{:} \StringTok{"date{-}time"}
    \FunctionTok{\},}
    \DataTypeTok{"requested\_model"}\FunctionTok{:} \FunctionTok{\{}
      \DataTypeTok{"type"}\FunctionTok{:} \StringTok{"string"}
    \FunctionTok{\},}
    \DataTypeTok{"resolved\_model"}\FunctionTok{:} \FunctionTok{\{}
      \DataTypeTok{"type"}\FunctionTok{:} \StringTok{"string"}
    \FunctionTok{\},}
    \DataTypeTok{"model\_identifier\_type"}\FunctionTok{:} \FunctionTok{\{}
      \DataTypeTok{"type"}\FunctionTok{:} \StringTok{"string"}\FunctionTok{,}
      \DataTypeTok{"enum"}\FunctionTok{:} \OtherTok{[}
        \StringTok{"fixed"}\OtherTok{,}
        \StringTok{"moving\_alias"}\OtherTok{,}
        \StringTok{"router"}\OtherTok{,}
        \StringTok{"unknown"}
      \OtherTok{]}
    \FunctionTok{\},}
    \DataTypeTok{"service\_tier"}\FunctionTok{:} \FunctionTok{\{}
      \DataTypeTok{"type"}\FunctionTok{:} \StringTok{"object"}\FunctionTok{,}
      \DataTypeTok{"additionalProperties"}\FunctionTok{:} \KeywordTok{false}\FunctionTok{,}
      \DataTypeTok{"required"}\FunctionTok{:} \OtherTok{[}\StringTok{"effective"}\OtherTok{]}\FunctionTok{,}
      \DataTypeTok{"properties"}\FunctionTok{:} \FunctionTok{\{}
        \DataTypeTok{"requested"}\FunctionTok{:} \FunctionTok{\{} \DataTypeTok{"type"}\FunctionTok{:} \StringTok{"string"} \FunctionTok{\},}
        \DataTypeTok{"effective"}\FunctionTok{:} \FunctionTok{\{} \DataTypeTok{"type"}\FunctionTok{:} \StringTok{"string"} \FunctionTok{\},}
        \DataTypeTok{"change\_reason"}\FunctionTok{:} \FunctionTok{\{}
          \DataTypeTok{"type"}\FunctionTok{:} \StringTok{"string"}\FunctionTok{,}
          \DataTypeTok{"enum"}\FunctionTok{:} \OtherTok{[}
            \StringTok{"none"}\OtherTok{,}
            \StringTok{"capacity"}\OtherTok{,}
            \StringTok{"rate\_limit"}\OtherTok{,}
            \StringTok{"policy"}\OtherTok{,}
            \StringTok{"provider\_failure"}\OtherTok{,}
            \StringTok{"unknown"}\OtherTok{,}
            \StringTok{"redacted"}
          \OtherTok{]}
        \FunctionTok{\}}
      \FunctionTok{\}}
    \FunctionTok{\},}
    \DataTypeTok{"effort"}\FunctionTok{:} \FunctionTok{\{}
      \DataTypeTok{"type"}\FunctionTok{:} \StringTok{"object"}\FunctionTok{,}
      \DataTypeTok{"additionalProperties"}\FunctionTok{:} \KeywordTok{false}\FunctionTok{,}
      \DataTypeTok{"required"}\FunctionTok{:} \OtherTok{[}\StringTok{"effective\_status"}\OtherTok{]}\FunctionTok{,}
      \DataTypeTok{"properties"}\FunctionTok{:} \FunctionTok{\{}
        \DataTypeTok{"requested"}\FunctionTok{:} \FunctionTok{\{}
          \DataTypeTok{"type"}\FunctionTok{:} \StringTok{"string"}\FunctionTok{,}
          \DataTypeTok{"enum"}\FunctionTok{:} \OtherTok{[}
            \StringTok{"minimal"}\OtherTok{,}
            \StringTok{"low"}\OtherTok{,}
            \StringTok{"medium"}\OtherTok{,}
            \StringTok{"high"}\OtherTok{,}
            \StringTok{"xhigh"}\OtherTok{,}
            \StringTok{"provider\_default"}\OtherTok{,}
            \StringTok{"unknown"}
          \OtherTok{]}
        \FunctionTok{\},}
        \DataTypeTok{"effective\_status"}\FunctionTok{:} \FunctionTok{\{}
          \DataTypeTok{"type"}\FunctionTok{:} \StringTok{"string"}\FunctionTok{,}
          \DataTypeTok{"enum"}\FunctionTok{:} \OtherTok{[}
            \StringTok{"completed"}\OtherTok{,}
            \StringTok{"budget\_exhausted"}\OtherTok{,}
            \StringTok{"downgraded"}\OtherTok{,}
            \StringTok{"not\_applicable"}\OtherTok{,}
            \StringTok{"unknown"}\OtherTok{,}
            \StringTok{"redacted"}
          \OtherTok{]}
        \FunctionTok{\}}
      \FunctionTok{\}}
    \FunctionTok{\},}
    \DataTypeTok{"tools"}\FunctionTok{:} \FunctionTok{\{}
      \DataTypeTok{"type"}\FunctionTok{:} \StringTok{"object"}\FunctionTok{,}
      \DataTypeTok{"additionalProperties"}\FunctionTok{:} \KeywordTok{false}\FunctionTok{,}
      \DataTypeTok{"required"}\FunctionTok{:} \OtherTok{[}\StringTok{"used"}\OtherTok{]}\FunctionTok{,}
      \DataTypeTok{"properties"}\FunctionTok{:} \FunctionTok{\{}
        \DataTypeTok{"allowed"}\FunctionTok{:} \FunctionTok{\{}
          \DataTypeTok{"type"}\FunctionTok{:} \StringTok{"array"}\FunctionTok{,}
          \DataTypeTok{"items"}\FunctionTok{:} \FunctionTok{\{} \DataTypeTok{"type"}\FunctionTok{:} \StringTok{"string"} \FunctionTok{\},}
          \DataTypeTok{"uniqueItems"}\FunctionTok{:} \KeywordTok{true}
        \FunctionTok{\},}
        \DataTypeTok{"used"}\FunctionTok{:} \FunctionTok{\{}
          \DataTypeTok{"type"}\FunctionTok{:} \StringTok{"array"}\FunctionTok{,}
          \DataTypeTok{"items"}\FunctionTok{:} \FunctionTok{\{} \DataTypeTok{"$ref"}\FunctionTok{:} \StringTok{"\#/$defs/tool\_use"} \FunctionTok{\}}
        \FunctionTok{\},}
        \DataTypeTok{"retrieval\_summary"}\FunctionTok{:} \FunctionTok{\{}
          \DataTypeTok{"type"}\FunctionTok{:} \StringTok{"object"}\FunctionTok{,}
          \DataTypeTok{"additionalProperties"}\FunctionTok{:} \KeywordTok{false}\FunctionTok{,}
          \DataTypeTok{"properties"}\FunctionTok{:} \FunctionTok{\{}
            \DataTypeTok{"source\_classes"}\FunctionTok{:} \FunctionTok{\{}
              \DataTypeTok{"type"}\FunctionTok{:} \StringTok{"array"}\FunctionTok{,}
              \DataTypeTok{"items"}\FunctionTok{:} \FunctionTok{\{} \DataTypeTok{"type"}\FunctionTok{:} \StringTok{"string"} \FunctionTok{\},}
              \DataTypeTok{"uniqueItems"}\FunctionTok{:} \KeywordTok{true}
            \FunctionTok{\},}
            \DataTypeTok{"retrieved\_item\_count"}\FunctionTok{:} \FunctionTok{\{}
              \DataTypeTok{"type"}\FunctionTok{:} \StringTok{"integer"}\FunctionTok{,}
              \DataTypeTok{"minimum"}\FunctionTok{:} \DecValTok{0}
            \FunctionTok{\},}
            \DataTypeTok{"redacted"}\FunctionTok{:} \FunctionTok{\{}
              \DataTypeTok{"type"}\FunctionTok{:} \StringTok{"boolean"}
            \FunctionTok{\}}
          \FunctionTok{\}}
        \FunctionTok{\}}
      \FunctionTok{\}}
    \FunctionTok{\},}
    \DataTypeTok{"context"}\FunctionTok{:} \FunctionTok{\{}
      \DataTypeTok{"type"}\FunctionTok{:} \StringTok{"object"}\FunctionTok{,}
      \DataTypeTok{"additionalProperties"}\FunctionTok{:} \KeywordTok{false}\FunctionTok{,}
      \DataTypeTok{"required"}\FunctionTok{:} \OtherTok{[}\StringTok{"input\_truncated"}\OtherTok{]}\FunctionTok{,}
      \DataTypeTok{"properties"}\FunctionTok{:} \FunctionTok{\{}
        \DataTypeTok{"input\_truncated"}\FunctionTok{:} \FunctionTok{\{}
          \DataTypeTok{"type"}\FunctionTok{:} \StringTok{"string"}\FunctionTok{,}
          \DataTypeTok{"enum"}\FunctionTok{:} \OtherTok{[}
            \StringTok{"false"}\OtherTok{,}
            \StringTok{"true"}\OtherTok{,}
            \StringTok{"unknown"}\OtherTok{,}
            \StringTok{"redacted"}
          \OtherTok{]}
        \FunctionTok{\},}
        \DataTypeTok{"retrieved\_item\_count"}\FunctionTok{:} \FunctionTok{\{}
          \DataTypeTok{"type"}\FunctionTok{:} \StringTok{"integer"}\FunctionTok{,}
          \DataTypeTok{"minimum"}\FunctionTok{:} \DecValTok{0}
        \FunctionTok{\},}
        \DataTypeTok{"context\_window\_class"}\FunctionTok{:} \FunctionTok{\{}
          \DataTypeTok{"type"}\FunctionTok{:} \StringTok{"string"}\FunctionTok{,}
          \DataTypeTok{"enum"}\FunctionTok{:} \OtherTok{[}
            \StringTok{"within\_limit"}\OtherTok{,}
            \StringTok{"near\_limit"}\OtherTok{,}
            \StringTok{"exceeded"}\OtherTok{,}
            \StringTok{"unknown"}\OtherTok{,}
            \StringTok{"redacted"}
          \OtherTok{]}
        \FunctionTok{\}}
      \FunctionTok{\}}
    \FunctionTok{\},}
    \DataTypeTok{"fallback"}\FunctionTok{:} \FunctionTok{\{}
      \DataTypeTok{"type"}\FunctionTok{:} \StringTok{"object"}\FunctionTok{,}
      \DataTypeTok{"additionalProperties"}\FunctionTok{:} \KeywordTok{false}\FunctionTok{,}
      \DataTypeTok{"required"}\FunctionTok{:} \OtherTok{[}\StringTok{"status"}\OtherTok{]}\FunctionTok{,}
      \DataTypeTok{"properties"}\FunctionTok{:} \FunctionTok{\{}
        \DataTypeTok{"status"}\FunctionTok{:} \FunctionTok{\{}
          \DataTypeTok{"type"}\FunctionTok{:} \StringTok{"string"}\FunctionTok{,}
          \DataTypeTok{"enum"}\FunctionTok{:} \OtherTok{[}
            \StringTok{"none"}\OtherTok{,}
            \StringTok{"occurred"}\OtherTok{,}
            \StringTok{"unknown"}\OtherTok{,}
            \StringTok{"redacted"}
          \OtherTok{]}
        \FunctionTok{\},}
        \DataTypeTok{"from"}\FunctionTok{:} \FunctionTok{\{} \DataTypeTok{"type"}\FunctionTok{:} \StringTok{"string"} \FunctionTok{\},}
        \DataTypeTok{"to"}\FunctionTok{:} \FunctionTok{\{} \DataTypeTok{"type"}\FunctionTok{:} \StringTok{"string"} \FunctionTok{\},}
        \DataTypeTok{"reason"}\FunctionTok{:} \FunctionTok{\{}
          \DataTypeTok{"type"}\FunctionTok{:} \StringTok{"string"}\FunctionTok{,}
          \DataTypeTok{"enum"}\FunctionTok{:} \OtherTok{[}
            \StringTok{"none"}\OtherTok{,}
            \StringTok{"rate\_limit"}\OtherTok{,}
            \StringTok{"provider\_error"}\OtherTok{,}
            \StringTok{"moderation\_refusal"}\OtherTok{,}
            \StringTok{"capacity"}\OtherTok{,}
            \StringTok{"policy"}\OtherTok{,}
            \StringTok{"unknown"}\OtherTok{,}
            \StringTok{"redacted"}
          \OtherTok{]}
        \FunctionTok{\}}
      \FunctionTok{\}}
    \FunctionTok{\},}
    \DataTypeTok{"safety"}\FunctionTok{:} \FunctionTok{\{}
      \DataTypeTok{"type"}\FunctionTok{:} \StringTok{"object"}\FunctionTok{,}
      \DataTypeTok{"additionalProperties"}\FunctionTok{:} \KeywordTok{false}\FunctionTok{,}
      \DataTypeTok{"required"}\FunctionTok{:} \OtherTok{[}\StringTok{"status"}\OtherTok{]}\FunctionTok{,}
      \DataTypeTok{"properties"}\FunctionTok{:} \FunctionTok{\{}
        \DataTypeTok{"status"}\FunctionTok{:} \FunctionTok{\{}
          \DataTypeTok{"type"}\FunctionTok{:} \StringTok{"string"}\FunctionTok{,}
          \DataTypeTok{"enum"}\FunctionTok{:} \OtherTok{[}
            \StringTok{"none"}\OtherTok{,}
            \StringTok{"intervened"}\OtherTok{,}
            \StringTok{"unknown"}\OtherTok{,}
            \StringTok{"redacted"}
          \OtherTok{]}
        \FunctionTok{\},}
        \DataTypeTok{"category"}\FunctionTok{:} \FunctionTok{\{} \DataTypeTok{"type"}\FunctionTok{:} \StringTok{"string"} \FunctionTok{\},}
        \DataTypeTok{"visible\_action"}\FunctionTok{:} \FunctionTok{\{}
          \DataTypeTok{"type"}\FunctionTok{:} \StringTok{"string"}\FunctionTok{,}
          \DataTypeTok{"enum"}\FunctionTok{:} \OtherTok{[}
            \StringTok{"none"}\OtherTok{,}
            \StringTok{"blocked"}\OtherTok{,}
            \StringTok{"masked"}\OtherTok{,}
            \StringTok{"rewritten"}\OtherTok{,}
            \StringTok{"refused"}\OtherTok{,}
            \StringTok{"unknown"}\OtherTok{,}
            \StringTok{"redacted"}
          \OtherTok{]}
        \FunctionTok{\}}
      \FunctionTok{\}}
    \FunctionTok{\},}
    \DataTypeTok{"region\_class"}\FunctionTok{:} \FunctionTok{\{}
      \DataTypeTok{"type"}\FunctionTok{:} \StringTok{"string"}\FunctionTok{,}
      \DataTypeTok{"enum"}\FunctionTok{:} \OtherTok{[}
        \StringTok{"user\_selected\_region"}\OtherTok{,}
        \StringTok{"data\_zone"}\OtherTok{,}
        \StringTok{"global"}\OtherTok{,}
        \StringTok{"provider\_default"}\OtherTok{,}
        \StringTok{"unknown"}\OtherTok{,}
        \StringTok{"redacted"}
      \OtherTok{]}
    \FunctionTok{\},}
    \DataTypeTok{"provider\_chain"}\FunctionTok{:} \FunctionTok{\{}
      \DataTypeTok{"type"}\FunctionTok{:} \StringTok{"array"}\FunctionTok{,}
      \DataTypeTok{"items"}\FunctionTok{:} \FunctionTok{\{} \DataTypeTok{"$ref"}\FunctionTok{:} \StringTok{"\#/$defs/provider\_hop"} \FunctionTok{\}}
    \FunctionTok{\},}
    \DataTypeTok{"completion\_status"}\FunctionTok{:} \FunctionTok{\{}
      \DataTypeTok{"type"}\FunctionTok{:} \StringTok{"string"}\FunctionTok{,}
      \DataTypeTok{"enum"}\FunctionTok{:} \OtherTok{[}
        \StringTok{"complete"}\OtherTok{,}
        \StringTok{"length\_limit"}\OtherTok{,}
        \StringTok{"tool\_error"}\OtherTok{,}
        \StringTok{"safety\_block"}\OtherTok{,}
        \StringTok{"error"}\OtherTok{,}
        \StringTok{"unknown"}
      \OtherTok{]}
    \FunctionTok{\},}
    \DataTypeTok{"redactions"}\FunctionTok{:} \FunctionTok{\{}
      \DataTypeTok{"type"}\FunctionTok{:} \StringTok{"array"}\FunctionTok{,}
      \DataTypeTok{"items"}\FunctionTok{:} \FunctionTok{\{} \DataTypeTok{"$ref"}\FunctionTok{:} \StringTok{"\#/$defs/redaction"} \FunctionTok{\}}
    \FunctionTok{\},}
    \DataTypeTok{"retention\_class"}\FunctionTok{:} \FunctionTok{\{}
      \DataTypeTok{"type"}\FunctionTok{:} \StringTok{"string"}\FunctionTok{,}
      \DataTypeTok{"enum"}\FunctionTok{:} \OtherTok{[}
        \StringTok{"ephemeral"}\OtherTok{,}
        \StringTok{"standard"}\OtherTok{,}
        \StringTok{"regulated"}\OtherTok{,}
        \StringTok{"audit\_hold"}\OtherTok{,}
        \StringTok{"unknown"}
      \OtherTok{]}
    \FunctionTok{\},}
    \DataTypeTok{"provider\_extensions"}\FunctionTok{:} \FunctionTok{\{}
      \DataTypeTok{"type"}\FunctionTok{:} \StringTok{"object"}
    \FunctionTok{\}}
  \FunctionTok{\},}
  \DataTypeTok{"$defs"}\FunctionTok{:} \FunctionTok{\{}
    \DataTypeTok{"tool\_use"}\FunctionTok{:} \FunctionTok{\{}
      \DataTypeTok{"type"}\FunctionTok{:} \StringTok{"object"}\FunctionTok{,}
      \DataTypeTok{"additionalProperties"}\FunctionTok{:} \KeywordTok{false}\FunctionTok{,}
      \DataTypeTok{"required"}\FunctionTok{:} \OtherTok{[}\StringTok{"name"}\OtherTok{,} \StringTok{"invocation\_count"}\OtherTok{]}\FunctionTok{,}
      \DataTypeTok{"properties"}\FunctionTok{:} \FunctionTok{\{}
        \DataTypeTok{"name"}\FunctionTok{:} \FunctionTok{\{} \DataTypeTok{"type"}\FunctionTok{:} \StringTok{"string"} \FunctionTok{\},}
        \DataTypeTok{"invocation\_count"}\FunctionTok{:} \FunctionTok{\{}
          \DataTypeTok{"type"}\FunctionTok{:} \StringTok{"integer"}\FunctionTok{,}
          \DataTypeTok{"minimum"}\FunctionTok{:} \DecValTok{0}
        \FunctionTok{\},}
        \DataTypeTok{"result\_refs"}\FunctionTok{:} \FunctionTok{\{}
          \DataTypeTok{"type"}\FunctionTok{:} \StringTok{"array"}\FunctionTok{,}
          \DataTypeTok{"items"}\FunctionTok{:} \FunctionTok{\{} \DataTypeTok{"type"}\FunctionTok{:} \StringTok{"string"} \FunctionTok{\}}
        \FunctionTok{\},}
        \DataTypeTok{"redacted"}\FunctionTok{:} \FunctionTok{\{} \DataTypeTok{"type"}\FunctionTok{:} \StringTok{"boolean"} \FunctionTok{\}}
      \FunctionTok{\}}
    \FunctionTok{\},}
    \DataTypeTok{"provider\_hop"}\FunctionTok{:} \FunctionTok{\{}
      \DataTypeTok{"type"}\FunctionTok{:} \StringTok{"object"}\FunctionTok{,}
      \DataTypeTok{"additionalProperties"}\FunctionTok{:} \KeywordTok{false}\FunctionTok{,}
      \DataTypeTok{"required"}\FunctionTok{:} \OtherTok{[}\StringTok{"role"}\OtherTok{]}\FunctionTok{,}
      \DataTypeTok{"properties"}\FunctionTok{:} \FunctionTok{\{}
        \DataTypeTok{"role"}\FunctionTok{:} \FunctionTok{\{}
          \DataTypeTok{"type"}\FunctionTok{:} \StringTok{"string"}\FunctionTok{,}
          \DataTypeTok{"enum"}\FunctionTok{:} \OtherTok{[}
            \StringTok{"requested"}\OtherTok{,}
            \StringTok{"served"}\OtherTok{,}
            \StringTok{"fallback"}\OtherTok{,}
            \StringTok{"tool"}\OtherTok{,}
            \StringTok{"unknown"}
          \OtherTok{]}
        \FunctionTok{\},}
        \DataTypeTok{"provider"}\FunctionTok{:} \FunctionTok{\{} \DataTypeTok{"type"}\FunctionTok{:} \StringTok{"string"} \FunctionTok{\},}
        \DataTypeTok{"model"}\FunctionTok{:} \FunctionTok{\{} \DataTypeTok{"type"}\FunctionTok{:} \StringTok{"string"} \FunctionTok{\},}
        \DataTypeTok{"redacted"}\FunctionTok{:} \FunctionTok{\{} \DataTypeTok{"type"}\FunctionTok{:} \StringTok{"boolean"} \FunctionTok{\}}
      \FunctionTok{\}}
    \FunctionTok{\},}
    \DataTypeTok{"redaction"}\FunctionTok{:} \FunctionTok{\{}
      \DataTypeTok{"type"}\FunctionTok{:} \StringTok{"object"}\FunctionTok{,}
      \DataTypeTok{"additionalProperties"}\FunctionTok{:} \KeywordTok{false}\FunctionTok{,}
      \DataTypeTok{"required"}\FunctionTok{:} \OtherTok{[}\StringTok{"field"}\OtherTok{,} \StringTok{"reason"}\OtherTok{]}\FunctionTok{,}
      \DataTypeTok{"properties"}\FunctionTok{:} \FunctionTok{\{}
        \DataTypeTok{"field"}\FunctionTok{:} \FunctionTok{\{} \DataTypeTok{"type"}\FunctionTok{:} \StringTok{"string"} \FunctionTok{\},}
        \DataTypeTok{"reason"}\FunctionTok{:} \FunctionTok{\{}
          \DataTypeTok{"type"}\FunctionTok{:} \StringTok{"string"}\FunctionTok{,}
          \DataTypeTok{"enum"}\FunctionTok{:} \OtherTok{[}
            \StringTok{"privacy"}\OtherTok{,}
            \StringTok{"security"}\OtherTok{,}
            \StringTok{"safety"}\OtherTok{,}
            \StringTok{"trade\_secret"}\OtherTok{,}
            \StringTok{"contractual"}\OtherTok{,}
            \StringTok{"not\_collected"}\OtherTok{,}
            \StringTok{"not\_applicable"}
          \OtherTok{]}
        \FunctionTok{\},}
        \DataTypeTok{"visible\_to"}\FunctionTok{:} \FunctionTok{\{}
          \DataTypeTok{"type"}\FunctionTok{:} \StringTok{"array"}\FunctionTok{,}
          \DataTypeTok{"items"}\FunctionTok{:} \FunctionTok{\{}
            \DataTypeTok{"type"}\FunctionTok{:} \StringTok{"string"}\FunctionTok{,}
            \DataTypeTok{"enum"}\FunctionTok{:} \OtherTok{[}
              \StringTok{"end\_user"}\OtherTok{,}
              \StringTok{"developer"}\OtherTok{,}
              \StringTok{"administrator"}\OtherTok{,}
              \StringTok{"auditor"}
            \OtherTok{]}
          \FunctionTok{\},}
          \DataTypeTok{"uniqueItems"}\FunctionTok{:} \KeywordTok{true}
        \FunctionTok{\}}
      \FunctionTok{\}}
    \FunctionTok{\}}
  \FunctionTok{\}}
\FunctionTok{\}}
\end{Highlighting}
\end{Shaded}

\end{document}